# Analysing the Effect of Masking Length Distribution of MLM: An Evaluation Framework and Case Study on Chinese MRC Datasets

Changchang. Zeng,[1,2] and Shaobo. Li,[3,1,2]

[1] Chengdu Institute of Computer Application, Chinese Academy of Sciences, Chengdu 610041, China.
[2] University of Chinese Academy of Sciences, Beijing 100049, China.
[3] State Key Laboratory of Public Big Data, Guizhou University, Guiyang 550025, China;

## Abstract

Machine reading comprehension (MRC) is a challenging natural language processing (NLP) task. Recently, the emergence of pre-trained models (PTM) has brought this research field into a new era, in which the training objective plays a key role. The masked language model (MLM) is a self-supervised training objective that widely used in various PTMs. With the development of training objectives, many variants of MLM have been proposed, such as whole word masking, entity masking, phrase masking, span masking, and so on. In different MLM, the length of the masked tokens is different. Similarly, in different machine reading comprehension tasks, the length of the answer is also different, and the answer is often a word, phrase, or sentence. Thus, in MRC tasks with different answer lengths, whether the length of MLM is related to performance is a question worth studying. If this hypothesis is true, it can guide us how to pre-train the MLM model with a relatively suitable mask length distribution for MRC task.

In this paper, we try to uncover how much of MLM's success in the machine reading comprehension tasks comes from the correlation between masking length distribution and answer length in MRC dataset. In order to address this issue, herein, (1) we propose four MRC tasks with different answer length distributions, namely short span extraction task, long span extraction task, short multiple-choice cloze task, long multiple-choice cloze task; (2) four Chinese MRC datasets are created for these tasks; (3) we also have pre-trained four masked language models according to the answer length distributions of these datasets; (4) ablation experiments are conducted on the datasets to verify our hypothesis. The experimental results demonstrate that our hypothesis is true.

## Introduction

In the field of natural language processing (NLP), machine reading comprehension (MRC) is a challenging task and has received extensive attention. According to the definition of Burges (2019), machine reading comprehension refers to: "A machine comprehends a passage of text if, for any question regarding that text that can be answered correctly by a majority of native speakers, that machine can provide a string which those speakers would agree both answers that question, and does not contain information irrelevant to that question (Burges, 2019) [1]."

Generally, MRC tasks can be roughly divided into four categories based on the answer form: cloze tests, multiple choice, span extraction and free answering [2,3]. Most of the early reading comprehension systems were based on retrieval technology, that is, search in the article according to the questions and find the relevant sentences as the answers. However, information retrieval mainly depends on keyword matching, and in many cases, the answers found by relying solely on text matching are not related to the questions.

With the development of machine learning (especially deep learning), and the release of large-scale datasets, the efficiency and quality of MRC model have been greatly improved. In some benchmark datasets, the accuracy of MRC model has exceeded the human performance [4]. In recent years, pre-trained language models (PTM) has brought revolutionary changes to the field of MRC. Among them, the most representative pre-trained model is the BERT proposed by Google in 2018 [5]. BERT uses unsupervised learning to pre-train on large-scale corpus, and creatively uses MLM and NSP subtasks to enhance the language ability of the model [5]. After the author released the code and pre-trained models, BERT was immediately used by researchers in various NLP tasks, and the previous SOTA results were refreshed frequently and significantly.

Recently, many efforts have been devoted to improve pre-trained models, and various pre-trained models have been proposed, such as: BERT-wwm [6], ERNIE 1.0 [7], ERNIE 2.0 [8], SpanBERT [9], MacBERT [10]. We can see that all of them have improved the masked language model (MLM) of the BERT model in different ways. However, the BERT model itself (the paradigm of the pre-training process, transformer based model and fine-tuning process) has not been significantly modified. This shows the importance of MLM. MLM is a self-supervised training objective of predicting missing tokens in a sequence from placeholders, which is widely used in various PTMs [11]. With the development of training objectives, many variants of MLM have been proposed, such as whole word masking [6], entity masking [7,8], phrase masking [7,8], and span masking [9].

In different MRC tasks, the length of the answer text is often different, and the answer is either a word, phrase, or sentence. Similarly, in different variants of MLM, the text length of the mask is also different. For example, the whole word masking improves the MLM objective of BERT by using the whole word instead of word piece [6]; the span masking performs the replacement at the span level and not for each token individually [9]; the entity masking masks entities that are usually composed of multiple words, while the phrase masking masks an entire phrase composed of multiple words as a conceptual unit [7,8].

How to choose masking scheme for MRC tasks with different answer lengths has become a question worth studying. At the same time, it also makes us wonder whether the length of MLM is related to their performance in MRC tasks with different answer lengths. If this hypothesis is true, maybe it can guide us how to pre-train an MLM model with a relatively suitable mask length distribution for various MRC tasks.

However, for different variants of MLM, there are many inconsistencies in their corpora, training methods, evaluation tasks and benchmark datasets. Therefore, it is difficult to perform ablation experiments on the existing MRC datasets and these publicly released pre-trained models to quantitatively measure the performance improvements brought about by different masking schemes.

To address the above issues, we design a set of controlled experiments to verify our hypothesis. In summary, our main contributions are as follows:

(1) Four MRC tasks with different answer length distributions are proposed, including short span extraction task, long span extraction task, short cloze task, long cloze task.

(2) We create MRC datasets for these four tasks, and statistically analyse the answer word length distribution on these four datasets;

(3) Using uniform hyper-parameters, we trained MLM with different masking length distributions.

(4) We conducted ablation experiments on the above dataset to verify our hypothesis. The experiment result shows that the consistency of the masking length distribution and the answers length distribution does affect the performance of the model, but it is not very significant, which indicates that there are other determinants besides the mask length of MLM.

## Related Works

**Existing Masked Language Models (MLMs)**

Recently, many efforts have been devoted to improve masked language models (MLM). In this section, we briefly introduce several existing MLMs, including word piece masking, whole word masking, entity masking, phrase masking, span masking, n-gram masking and so on.

**Word Piece Masking.** Word piece masking is the MLM used in the original version of BERT [5], where the 'WordPiece Tokenizer' is used in the data pre-processing to spilt the input sequence into sub-words, which is very effective in dealing with out of vocabulary (OOV) words. In Chinese text tokenization, when a sentence is tokenized with a 'WordPiece Tokenizer', it will be split into Chinese characters. Then, tokens are selects randomly for masking, 15% of the tokens will be randomly selected. Among the selected tokens, for each word, it has an 80% probability of being replaced with [MASK], 10% will be replaced with a random token, and 10% will remain unchanged. It should be noted that each token is masked independently according to the above probability, rather than all selected tokens are masked at the same time.

**Whole Word Masking.** Whole word masking (WWM) is an upgraded version of BERT [5] released by Google, which mitigate the drawbacks of masking partial WordPiece tokens in original BERT [6]. In the whole word masking, if a subword of a complete word is masked, the other parts of the same word will also be masked, that is, the whole word will be masked at the same time. In the Chinese version of BERT released by Google, Chinese is segmented at the granularity of characters, and the Chinese word segmentation (CWS) [12] is not considered. Therefore, Cui et al. (2019) applied the whole word masking to Chinese [6], and masked the whole word instead of masking Chinese characters.

**Entity Level Masking.** Entities usually contain important information in the sentences, such as a person, location, organization, product, etc. Unlike selecting random tokens for masking, entity level masking masks the whole named entities, which are usually composed of multiple words [7,8]. Before masking, the text needs to be segmented using named entity recognition tools. In entity level masking, the MLM implicitly learned the information about longer semantic dependency, such as the relationship between entities.

**Phrase Level Masking.** A phrase is a small group of words or characters as a conceptual unit. Phrase level masking mask the whole phrase which is composed of several words [7,8], and it is similar to the N-gram masking scheme [10,5,9]. For English, vocabulary analysis and chunking tools are we used to get the boundaries of phrases in sentences, and use some language-related segmentation tools to get word/phrase information in other languages (such as Chinese) [7,8]. In this way, the prior knowledge of phrases is considered to be learned implicitly during the training procedure, such as syntactic and semantic information.

**Span Masking.** Span masking was proposed in the SpanBERT [9], in which contiguous random spans are masked, rather than individual tokens. The process of span masking is, first, iteratively select contiguous random spans until the 15% masking budget is spent. In each iteration, the length of span is selected according to the geometric distribution. Then, randomly select the starting point of the span. Third, replace all the tokens in the same selected span with [MASK], random or original tags according to the 80%-10%-10% rule in BERT [5], where the span constitutes the unit. Therefore, it forces the model to use only the context in which the span occurs to predict the entire span.

**N-gram Masking.** N-gram masking is usually considered to be first proposed by Devlin et al. (2019), according to their model name on the SQuAD leaderboard [10]. In N-gram masking, a sequence of N words is treated as a whole unity. During pre-training of MLM, all words in the same unit are masked, instead of masking only one word or character. N-gram masking is used in many advanced pre-training models. For example, MacBERT [10] uses N-gram masking scheme for selecting candidate tokens for masking, with a percentage of 40%, 30%, 20%, and 10% for word-level unigram to 4-gram. To a certain extent, the span masking, entity level masking and phrase level masking can be regarded as special cases of N-gram masking scheme [5,9].

**Explicitly N-gram Masking.** Explicitly N-gram masking is an explicit N-gram masking scheme, in which N-gram are replaced by a single [MASK] symbol [13]. When predicting masked tokens, explicit N-gram identities are directly used instead of token sequences. In addition, Explicitly N-gram masking uses a generator model to sample reasonable N-gram identities as an optional N-gram mask, and predicts it in both coarse-grained and fine-grained manner to achieve comprehensive relation modelling. Explicitly N-gram masking is proposed by Baidu team in 2021 [13].

**Multi-level Masking.** The multi-level masking is a uses multiple scheme at the same time. For example, the knowledge masking proposed in the ERNIE 1.0 [7] masking can be regarded as a kind of multi-level masking scheme, which uses both phrase-level masking and entity-level masking. The knowledge masking treats a phrase or entity as a unit, which is usually composed of several words. All words in the same unit are masked, instead of masking only one word or character. The knowledge masking does not directly add knowledge embedding, but is considered to be learning information about knowledge, such as entity attributes and event types, to guide word embedding learning [7].

**Dynamic Masking.** Static masking is used in the MLM of the original BERT, and the masking is performed only once during the data pre-processing before MLM training, which means that the same words are masked in the input sequence provided to the model on each epoch. In order to avoid masking the same words multiple times and make full use of the input sequence, dynamic masking is proposed. In the dynamic masking process, the "dupe_factor" is defined, and the input sequence will be duplicated "dupe_factor", then the same sequence will have different masks [14]. Before providing input sequence to the model each time, the masking operation will be performed repeatedly. Therefore, the model will see different masking versions of the same sequence. Dynamic masking is adopted by many pre-trained models, such as RoBERTa [14].

**Interpretability of Masked Language Models**

With various advanced MLMs, many pre-trained language models have achieved the state-of-the-art performance when adapted to MRC task. The black box nature of MLM and related pre-trained models has inspired many works trying to understand them.

Many efforts have been devoted to uncover whether the MLM calculates various types of structured information by probing analysis, or evaluating the performance of simple classifiers on the representations [15,16,17,18,19,20]. Popular methods also include analysing self-attention weights, and evaluating the performance of classifiers on with different representations as inputs [21]. A possible explanation for the success of masked language model (MLM) training is that these models have learned to represent the semantic information or syntactic information [22].

**Semantic Information.** With MLM probing study, Ettinger et al. (2020) applied a set of diagnostic methods derived from human language experiments to the BERT model and found that BERT has a certain understanding of semantic roles [21,23].

Tenney et al. (2019) used a set of detection tasks derived from traditional NLP pipelines to quantify the encoding position of specific types of Semantic information, and The experimental results show that BERT encodes information about entity types, relationships, semantic roles, and prototype roles [21,24].

**Syntactic Information.** Through the probes of MLM, Goldberg assessed the extent to which the BERT model captures English syntactic phenomena and found the BERT models perform remarkably well on the syntactic test cases. The experimental results show that BERT considers subject-predicate agreement when completing the cloze task, even for meaningless sentences and sentences with participle clauses between subject and verb [21,25].

Wu et al. (2020) proposed a perturbation masking technique to evaluate the impact of one word on the prediction of another word in MLM. They concluded that BERT ''naturally'' learns some syntactic information, although it is not very similar to linguistic annotated resources [21,26].

**Distributional Information.** Most recently, Sinha et al. (2021) surprisingly found that most of MLM's high performance can in fact be explained by the "distributional prior" rather than its ability to replicate "the types of syntactic and semantic abstractions traditionally believed necessary for language processing" (Tenney et al., 2019) [24]. In other words, they found that the success of MLM in downstream tasks is almost entirely because they can model high-order word co-occurrence statistics. To prove this, they pre-trained MLMs on sentences with random shuffled word order, and showed that after fine-tuning many downstream tasks, these models can still achieve high accuracy, including on tasks designed specifically to be challenging for models that ignore word order. According to some parametric syntactic probes, these models perform surprisingly well, which indicates possible deficiencies when testing the representation for syntactic information [22].

**Motivation and Approach**

Firstly, as described in section 2.1, most of the existing MLMs adopt different mask rules to improve their performance. Some take a word as a mask unit, some take phrases, entities or spans as mask units, and others adopt multi-level masking schemes. As we can see, the length of masked text is one of the basic variables in the above masking schemes. However, at present, there is a lack of quantitative research and analysis on the performance of MLM with different masking lengths. The whole word masking [6] only masks words, and entity and phrase masking schemes [7,8] mask only entities or phrases. Span masking [9] simply uses geometric distribution in the process of selecting span. N-gram masking in MacBERT [10] also just masks different N-grams in a fixed probability. However, for MRC tasks with different answer lengths, if different masking lengths of MLM are used. Will the performance achieved be different? There is no relevant research yet. In addition, some MLMs use a multi-level masking scheme, such as the knowledge masking in ERINE [7,8]. However, what is the optimal proportion of masking schemes of different levels? It is also a question worth studying. If there is a correlation between the performances of MLM and masking lengths in MRC tasks with different answer lengths, then, we can choose an appropriate length for an MLM according to the length distribution in the MRC dataset.

Secondly, from the interpretability of MLM described in section 2.2, we can see that the theoretical analysis of the MLM is very challenging. There are many empirical studies trying to understand why MLMs are so effective, and one possible explanation for the impressive performance of MLMs is that these models have learned the semantic information and syntactic information. A lot of work has been devoted to revealing whether MLM calculates various types of structured information [15,16,17,18,19,20,21,23,24,25,26].

However, the most recent studies have pointed out that the success of MLM may actually come from the word distribution information it learns to a large extent, and they found that the success of MLM in downstream tasks is almost entirely because they can model high-order word co-occurrence statistics [22].

Inspired by this research, we wonder whether the distribution of masking length will also affect the performance of MLM in MRC tasks with different answer length. In this work, we try to uncover how much of MLM's success comes from the correlation between masking length distribution and answer length in MRC dataset. We treat the distribution of answer lengths in the MRC dataset and the masking length of MLMs as latent variables, and treat the performance of different MLMs on downstream MRC tasks with different answer distributions as functions of latent variables. Assuming that the distribution of the answer length in the MRC dataset is correlated with the masking length of the MLM. Then pre-training on a large corpus allows MLM to learn the hidden information of different length. Therefore, in the downstream MRC task, the MLM model whose masking length is closest to the answer length distribution in the MRC dataset should achieve better performance.

The key start point of our research work is to propose an evaluation framework to quantitatively verify whether masking schemes of different lengths will affect the results of the MLM language model in MRC tasks with different answer length. However, using the existing pre-trained models and MRC dataset to create a verification framework is challenging because there are too many different factors affecting performance. Since there are many inconsistencies in the pre-training corpus, pre-training methods, downstream tasks, and evaluation datasets used by different pre-trained models. Therefore, it is difficult to conduct ablation experiments on different masking schemes.

To address the above issues, we first design four MRC tasks and construct the related datasets with different answer lengths. Next, using unified hyper-parameters, we retrain several MLMs with different masking lengths according to the answer lengths of the above MRC datasets. Then, ablation experiments are carried out, and we evaluate the performance of different MLMs on the above MRC dataset. The key points of our experiment are as follows:

(1) New MRC Tasks and Datasets

When designing the MRC tasks and datasets, we integrate the mainstream MRC tasks, namely cloze test, multiple choice, span extraction and free answering [2,3], and we adopt two kinds of MRC tasks, including the span extraction tasks and two multiple choice cloze tasks. The answer length of these two tasks can also be divided into two categories: long answer and short answer. Finally, we construct four MRC datasets, namely short span dataset, long span dataset, short cloze dataset, and long cloze dataset. In addition, because the Chinese corpus is composed of Chinese characters, there is no well-marked word boundary, which is conducive to eliminate the influence of word boundary information in the pre-trained model. Therefore, we chose to create Chinese MRC datasets.

(2) Training MLM from Scratch

Existing MLMs usually integrates a variety of improvements, such as MLM in the SpanBERT [9] which uses both span masking and the SBO pre-training tasks [9] at the same time. In order to eliminate the influence of prior knowledge embedded in the pre-trained model, in this experiment, we do not directly use the existing MLMs, but conducted MLM training from scratch by ourselves, thereby eliminating the interference variables.

(3) Unified Pre-training Corpora

When training MLMs with different masking lengths, we use the same pre-training corpora to eliminate the impact of word distribution in different corpora.

(4) Answer Length Distribution of Datasets

In this article, in order to quantitatively verify whether masking schemes of different lengths will affect the performance of the MLM language model, we have counted the length distributions of different datasets.

(5) Masking Length Distribution of MLMs

In the process of training different MLMs, we use the weighted average answer length distribution in the data set as the MLM mask length. To quantitatively verify whether masking schemes of different lengths will affect the results of the MLM language model.

(6) Unified Masking Ratios

During the experiment, we fixed the masking ratio to be the same as the original version of BERT. That is, select 15% of the text in the paragraph, and 80% of the selected text are replaced by [MASK], 10% are replaced by random tokens, and 10% are replaced by original tokens, 10% remain unchanged, and 10% replaced with random tokens. We perform this replacement at the sequence level, that is, each time all tokens in a sequence are replaced with "mask", or random tokens, or remain unchanged.

(7) Unified Pre-training Hyper-Parameters

In order to eliminate the influencing factors, we use the same model hyper-parameters in the pre-training of different MLMs.

## Proposed MRC Tasks

According to the style of the answers and questions, MRC tasks can be roughly divided into four categories: cloze test, multiple choice, span extraction and free answering. [2,3].When designing MRC tasks and datasets required for ablation experiments, we integrate the main characteristics of these MRC tasks, and we adopt two kinds of MRC tasks, including the span extraction tasks and two multiple-choice cloze tasks. The answer length of these two tasks can also be divided into two categories: long answer and short answer. Finally, the four MRC tasks are:

(1) Span extraction tasks with short answers;

(2) Span extraction tasks with long answers;

(3) Multiple-choice cloze tasks with short answers;

(4) Multiple-choice cloze tasks with long answers;

We believe these tasks are representative of most of the current MRC tasks. Among them, the number of tokens in the short answer of span extraction tasks is set to be greater than 3 and less than 7, and the size of the long answer is greater than 6 and less than 10; The number of tokens in the short answer of the multiple-choice cloze tasks is greater than 6 and less than 15, and the size of the long answer is greater than 16 and less than 30.

In the following subsections, we briefly introduce the definitions of typical MRC tasks and the two types of MRC tasks we used in the experiment.

**Typical MRC Tasks**

Generally, the definition of a typical MRC task is given below:

> **Definition 1.** Typical machine reading comprehension task could be formulated as a supervised learning problem. Given the training examples $\{p, q, a\}$, where $p$ is a passage, and $q$ is a question. The goal of typical machine reading comprehension task is to learn a predictor $f$ which takes the passage $p$ and a corresponding question $q$ as inputs and gives the answer $a$ as output, which could be formulated as the following formula [2,3,4]:

$$a = f(p, q) \tag{1}$$

and it is necessary that a majority of native speakers would agree that the question $q$ does regarding that text $p$, and the answer $a$ is a correct one which does not contain information irrelevant to that question.

**Span Extraction Tasks with Different Answer Lengths**

In order to quantitatively verify whether masking schemes with different lengths will affect the performance of MLM, we propose two span extraction tasks with different answer lengths for Chinese machine reading comprehension. Table 1 shows an example in the proposed span extraction task.

Table 1: An example in the proposed span extraction task.

| **Context**: 传染病的医学治疗属于传染病医学领域，在某些情况下传播学的研究属于**流行病学**领域。一般来说，感染最初由初级保健医生或内科专家诊断。例如，"简单"肺炎通常由内科医师或肺科医师（肺科医师）治疗。因此，传染病专家的工作需要与病人和全科医生以及实验室科学家、免疫学家、细菌学家和其他专家合作。 | **Context**: The medical treatment of infectious diseases belongs to the field of infectious disease medicine. In some cases, the research of communication belongs to **the field of epidemiology**. In general, infection is initially diagnosed by a primary care physician or medical expert. For example, "simple" pneumonia is usually treated by a physician or pulmonary physician (pulmonary physician). Therefore, the work of infectious disease experts requires cooperation with patients and general practitioners, as well as laboratory scientists, immunologists, bacteriologists and other experts. |
|---|---|
| **Question**: 疾病传播可以归入哪些研究领域的范畴？ | **Question**: What research areas can disease transmission fall into? |
| **Answer:** 流行病学 | **Answer:** the field of epidemiology |

Definition of the span extraction task is：

> **Definition 2.** Given a serial of training samples. Each sample contains a passage about a public service event, a corresponding question and the answer to this question. The answer should be a span which is directly extracted from the passage. The goal of Span Extraction machine reading comprehension task is to train the machine so that it can find the correct answers in the given passage. The task can be simplified by predicting the start and end pointer of the right answer in the given passage.

**Multiple-choice Cloze Tasks with Different Answer Lengths**

We also proposed two multiple-choice cloze tasks with different answer lengths for Chinese machine reading comprehension. The form of our multiple-choice cloze tasks is similar to the CMRC2019 task [27], but redundant fake answers are removed. Table 2 shows an example in the proposed multiple-choice cloze task.

Table 2: An example in the proposed multiple-choice cloze task.

| **Passage**: | **Passage**: |
|---|---|
| 2001年初，苹果开始销售带有CD-RW驱动器的电脑，并强调Mac电脑的DVD播放能力，将DVD-ROM和DVD-RAM驱动器作为标准配置。[BLANK1]，苹果在可写CD技术方面"迟到了"，但他认为，Mac电脑可以成为一个"数字中心"，连接并促成一种"新兴的数字生活方式"。苹果随后将对其iTunes音乐播放器软件进行更新，同时推出一项颇具争议的"翻录、混合、刻录"广告活动，[BLANK2]。[BLANK3]，iPod是苹果第一款成功的手持设备。[BLANK4]，例如失败的Power Macg4cube、面向教育的EMAC和面向专业人士的钛（后来的铝）PowerBook G4笔记本电脑。直到1989年苹果公司发布了他们的第一台便携式电脑，[BLANK5]。[BLANK6]，它很快在1991年被第一条PowerBook线取代：PowerBook100，[BLANK7]；16MHz 68030 PowerBook140；25MHz 68030 PowerBook170。[BLANK8]，在键盘前面有一个内置的指向设备（轨迹球）的便携式计算机。1993年的PowerBook165C是苹果第一台采用彩色屏幕的便携式电脑，[BLANK9]，分辨率为640x 400像素。 | In early 2001, Apple began selling computers with CD-RW drives, emphasized the DVD playback capability of MAC computers, and took DVD-ROM and DVD-RAM drives as standard configurations. [BLANK1] Apple was "late" in the technology of writable CD, but he believed that the MAC could become a "Digital Center" to connect and promote an "emerging digital lifestyle". Apple will then update its iTunes music player software, and launch a controversial "rip, mix and burn" advertising campaign,[BLANK2],[BLANK3], iPod is Apple's first successful handheld device[BLANK4], such as the failed Power Macg4cube, the EMAC for education, and the titanium (later aluminum) PowerBook G4 laptop for professionals. Until 1989, Apple released their first portable computer, [BLANK5]. [BLANK6], which was soon replaced by the first PowerBook line in 1991: powerbook100, [BLANK7]; 16MHz 68030 PowerBook140；25MHz 68030 PowerBook170。[BLANK8], a portable computer with a built-in pointing device (trackball) in front of the keyboard. The 1993 PowerBook 165C was Apple's first portable computer with a color screen, [BLANK9], with a resolution of 640x 400 pixels. |
| **Options**:<br>A:"显示256种颜色"<br>B:"尽管由于大量的设计问题"<br>C:"一些人认为这鼓励了媒体盗版"<br>D:"伴随着iPod的发布"<br>E:"使其能够刻录CD"<br>F:"一种小型便携式设备"<br>G:"史蒂夫•乔布斯承认"<br>H:"Macintosh便携式电脑"<br>I:"苹果继续推出产品" | **Options**:<br>A:"display 256 colors"<br>B:"despite a large number of design problems"<br>C:"some people think this encourages media piracy"<br>D:"with the release of iPod"<br>E:"enable it to burn CDs"<br>F:"a small portable device"<br>G:"Steve Jobs admitted that "<br>H:"Macintosh laptop"<br>I:"Apple continues to launch products" |
| **Answer:**<br>G, I, H, F, C, D, B, E, A | **Answer:**<br>G, I, H, F, C, D, B, E, A |

Definition of the multiple-choice cloze task is：

> **Definition 3.** Generally, the reading comprehension task can be described as a triple 〈P, Q, A〉, where P represents Passage, Q represents Question, and the A represents Answer. Specifically, for multiple-choice cloze-style reading comprehension task, we select several sentences in the passages and replace with special marks (for example, [BLANK]), forming an incomplete passage. The selected sentences form a candidate list, and the machine should fill in the blanks with these candidate sentences to form a complete passage [2,3,4,27].

## Evaluation Metrics

In this paper, we use F1 and EM to measure the performance of the pre-trained model in the span extraction tasks.

**F1 Score**

F1 is a commonly used MRC task evaluation metric. The equation of F1 for a single question is:

$$F1 = \frac{2 \times Precision \times Recall}{Precision + Recall} \qquad (2)$$

Where *P* denotes the token-level Precision for a single question and *R* denotes the Recall for a single question [2,3,4].

**Precision**

Precision represents the percentage of maximum span overlap between the tokens in the correct answer and the tokens in the predicted answer. In order to calculate Precision, we first need to obtain true positive (TP), false positive (FP), true negative (TN) and false negative (FN), as shown in Figure 1:

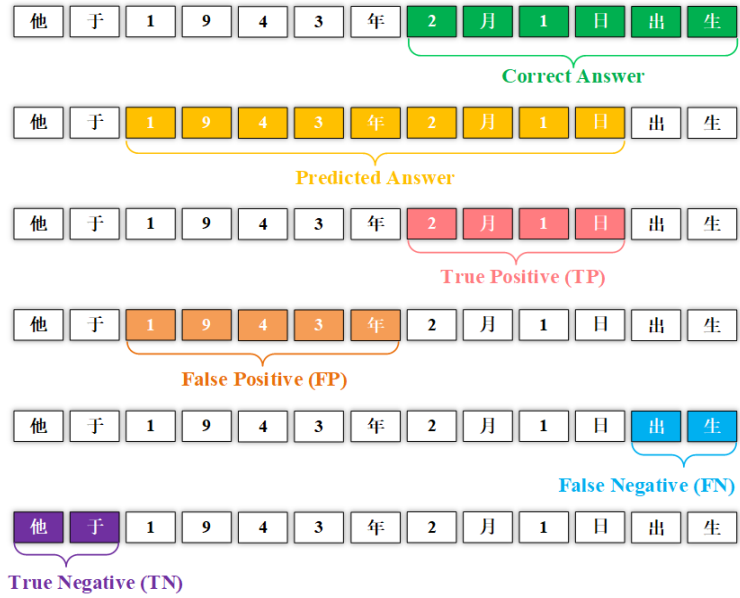

Figure 1: The true positive (TP), false positive (FP), true negative (TN), and false negative (FN).

As shown in Figure 1, for a single question in the proposed dataset, the true positive (TP) is equal to the maximum common span (MCS) between the predicted answer and the correct answer. False positive (FP) indicates the span not in the correct answer but in the predicted answer, while false negative (FN) indicates the span not in the predicted answer but in the correct answer [2,3,4]. The Precision of a single question is calculated as follows:

$$Precision = \frac{TP}{TP+FP} \qquad (3)$$

**Recall**

Recall represents the percentage of correct answers that have been correctly predicted in the question [2,3,4]. According to the above definitions of true positive (TP), false positive (FP) and false negative (FN), the Recall of a single answer is calculated as follows:

$$Recall = \frac{TP}{TP+FN} \quad (4)$$

Where Recall represents the recall rate of a single problem, NumPT represents the number of true positive (TP) tokens, and NumFN represents the number of false negative (FN) tokens.

**Exact Match**

Exact Match represents the percentage of questions where the answer generated by the system exactly matches the correct answer, which means that every word is the same. Exact match is usually abbreviated as EM. In the span extraction MRC task, the answer to the question is a sentence, and some words in the predicted answer may be included in the correct answer, while other words are not included in the correct answer [2,3,4]. For example, if the MRC task contains $N$ questions, each question corresponds to a correct answer. The answer can be a word, a phrase or a sentence, and the number of predicted answers exactly the same as the correct answer is $M$. Exact Match can be calculated as follows:

$$Exact\ Match = \frac{M}{N} \quad (5)$$

**Accuracy**

In this paper, we use Accuracy to measure the performance of the pre-trained model in the multiple choice cloze tasks. Accuracy is defined as the ratio of the number of correctly predicted samples to the total number of samples for a given test dataset.

For example, suppose a MRC task contains $N$ questions, each question corresponds to one correct answer, the answers can be a word, a phrases, or a sentence, and the number of questions that the system answers correctly is $M$. The equation for the accuracy is as follows:

$$Accuracy = \frac{M}{N} \quad (6)$$

In addition, in order to make the assessment more reliable, following the evaluation method of CMRC2019 [27], we adopt two metrics to evaluate the systems on our datasets, which are Question-level Accuracy (QAC) and Passage-level Accuracy (PAC).

The Question-level Accuracy (QAC) is the ratio between the correct prediction and the total blanks, which can be calculated by the following formula [27]:

$$QAC = \frac{\#\ Correctly\ Answered\ Questions}{\#\ Totlal\ Questions} \quad (7)$$

Similar to the QAC, Passage-level Accuracy (PAC) is to measure how many passages have been correctly answered. We only count the passages that all blanks have been correctly predicted [27].

Passage-level accuracy (PAC) is used to measure how many passages are answered exactly correctly. Similar to the Exact Match, only paragraphs that all blanks are correctly predicted are considered as exactly correctly predicted samples. Passage-level accuracy (PAC) can be calculated by the following formula [27]:

$$PAC = \frac{\# \text{ Exactly Correct Paragraphs}}{\# \text{ Totlal Paragraphs}} \quad (8)$$

## Dataset Construction

As mentioned above, in order to eliminate the influence of interference factors on the experiment as much as possible, we designed four MRC tasks: short span extraction task, long span extraction task, short multiple-choice cloze task and long multiple-choice cloze task. In this section, we further construct four Chinese MRC datasets for these MRC tasks. Unlike English text, a feature of Chinese text is that there are no obvious spaces to mark word boundaries, so the influence of word boundary information on the results can be further eliminated. So in this article, we use Chinese as the language of the dataset. Below, we will briefly introduce the construction methods of these MRC datasets.

**Span Extraction Dataset with Different Answer Lengths**

The corpus of our span extraction datasets comes from the paragraphs in the Chinese SQuAD dataset [28]. The Stanford Question Answering Dataset (SQuAD) [29] is one of the most popular machine reading comprehension datasets, containing more than 100,000 questions generated by human, and the answer to each question is a span of text in a related context [20]. Since its release in 2016, SQuAD 1.1 has quickly become the most widely used MRC dataset. Now it has been updated to SQuAD 2.0 [4,30].

The Chinese SQuAD dataset [28] is translated from the original SQuAD through machine translation and manual correction, including SQuAD 1.1 [29] and SQuAD 2.0 [30]. Because some translations cannot find the answers in the original text (the answer translation and document translation are different), the amount of data is reduced compared to the original English version of SQuAD. After data cleaning, the Chinese SQuAD dataset contains 125,892 questions, 36,100 paragraphs, and the number of unanswerable questions is 49,443 [28]. Among them, each paragraph includes a number of different contexts, and each context includes multiple question and answer pairs. Then, we divided the paragraphs in the Chinese SQuAD dataset according to the length of the answer, and obtained the long span extraction dataset and the short span extraction dataset, where the number of tokens in the short answer of span extraction tasks is set to be greater than 3 and less than 7, and the size of the long answer is greater than 6 and less than 10. The statistics of our span extraction datasets is shown in sections below.

**Multiple-choice Cloze Dataset with Different Answer Lengths**

The corpus source of our multiple-choice cloze dataset is the NLPCC2017 corpus [31]. The cleaned NLPCC2017 corpus contains 50,000 news articles with summary and the average number of tokens in an article is 1036 [31].

We first divide the above corpus into several paragraphs, and then divide each paragraph into sentences using commas, periods, semicolons, exclamation marks, and question marks as the dividing point. Then, when constructing the multiple-choice cloze dataset with short answers, for each paragraph, we randomly select 9 sentences as candidate long answers, and the number of tokens in these sentences is greater than 6 and less than 15. When constructing the multiple-choice cloze dataset with long answers, for each paragraph, we also randomly select 9 sentences as candidate long answers, and the number of tokens in these sentences is greater than 16 and less than 30. After selecting the candidate answers, we randomly shuffle the order of the answers to obtain candidate options in the form of multiple choices. The statistics of our multiple-choice cloze datasets is shown in sections below.

**Dataset Analysis**

In this subsection, we analyse the paragraphs, questions and answers in the proposed datasets. Specifically, we explore (1) the statistics of the data size, (2) the length distribution of the answer lengths in the train set, development set and test set of the proposed datasets. As we can see, the statistics of the proposed span extraction datasets are given in Table 3. Table 4 also shows the statistics of the proposed multiple-choice cloze datasets.

Table 3: Statistics of the proposed span extraction datasets.

|  | Short Span Extraction | | | Long Span Extraction | | |
| --- | --- | --- | --- | --- | --- | --- |
|  | **Train Set** | **Dev Set** | **Test Set** | **Train Set** | **Dev Set** | **Test Set** |
| Paragraph # | 600 | 150 | 200 | 600 | 150 | 200 |
| Passage # | 15,265 | 2,770 | 2,484 | 7,756 | 1,163 | 1,565 |
| Question # | 31,390 | 6,774 | 8,309 | 12,053 | 2,376 | 5,812 |
| Max Tokens in a Context # | 512 | 512 | 512 | 512 | 512 | 512 |
| Max Answer Tokens # | 6 | 6 | 6 | 9 | 9 | 9 |
| Min Answer Tokens # | 4 | 4 | 4 | 7 | 7 | 7 |

Table 4: Statistics of the proposed multiple-choice cloze datasets.

|  | Short Cloze | | | Long Cloze | | |
| --- | --- | --- | --- | --- | --- | --- |
|  | **Train Set** | **Dev Set** | **Test Set** | **Train Set** | **Dev Set** | **Test Set** |
| Passages # | 4,500 | 1,000 | 1,000 | 4,500 | 1,000 | 1,000 |
| Blanks # | 4,0500 | 9,000 | 9,000 | 4,0500 | 9,000 | 9,000 |
| Max Tokens in a Passage# | 1,000 | 1,000 | 1,000 | 1,000 | 1,000 | 1,000 |
| Max Answer Tokens # | 14 | 14 | 14 | 29 | 29 | 29 |
| Min Answer Tokens # | 7 | 7 | 7 | 17 | 17 | 17 |
| Options # | 9 | 9 | 9 | 9 | 9 | 9 |

**Distribution of Answer Length:**

We have separately counted the distribution of answer lengths in these four datasets. Table 5 and Table 6 show the answers length distributions of the train set, development set and test set. For example, in the short span extraction dataset, there are 16,171 answers that has 4 tokens in the training set, and 3,344 answers that has 4 tokens in the development set, while 4147 in the test set.

Table 5: The answers length distributions of the proposed span extraction datasets.

| | **Short Span Dataset** | | | | | |
| --- | --- | --- | --- | --- | --- | --- |
| # Tokens | #Train | PP % | # Dev | PP % | # Test | PP % |
| 4 | 16171 | 51.52% | 3344 | 49.37% | 4147 | 49.91% |
| 5 | 8566 | 27.29% | 1828 | 26.99% | 2230 | 26.84% |
| 6 | 6653 | 21.19% | 1602 | 23.65% | 1932 | 23.25% |
| Total | 31390 | 100.00% | 6774 | 100.00% | 8309 | 100.00% |
| | **Long Span Dataset** | | | | | |

| # Tokens | # Train | PP % | # Dev | PP % | # Test | PP % |
| --- | --- | --- | --- | --- | --- | --- |
| 7 | 5040 | 41.82% | 819 | 34.47% | 2071 | 35.63% |
| 8 | 4192 | 34.78% | 980 | 41.25% | 2390 | 41.12% |
| 9 | 2821 | 23.40% | 577 | 24.28% | 1351 | 23.25% |
| Total | 12053 | 100.00% | 2376 | 100.00% | 5812 | 100.00% |

Table 6: The answers length distributions of the proposed multiple-choice cloze datasets.

| Short Cloze Dataset | | | | | | |
| --- | --- | --- | --- | --- | --- | --- |
| # Tokens | #Train | PP % | # Dev | PP % | # Test | PP % |
| 7 | 4428 | 10.93% | 974 | 10.82% | 1016 | 11.29% |
| 8 | 4630 | 11.43% | 1022 | 11.36% | 1037 | 11.52% |
| 9 | 5020 | 12.40% | 1166 | 12.96% | 1077 | 11.97% |
| 10 | 5260 | 12.99% | 1212 | 13.47% | 1154 | 12.82% |
| 11 | 5389 | 13.31% | 1149 | 12.77% | 1253 | 13.92% |
| 12 | 5427 | 13.40% | 1205 | 13.39% | 1200 | 13.33% |
| 13 | 5308 | 13.11% | 1171 | 13.01% | 1136 | 12.62% |
| 14 | 5038 | 12.44% | 1101 | 12.23% | 1127 | 12.52% |
| Total | 40500 | 100.00% | 9000 | 100.00% | 9000 | 100.00% |
| Long Cloze Dataset | | | | | | |
| # Tokens | #Train | PP % | # Dev | PP % | # Test | PP % |
| 17 | 5209 | 12.86% | 1132 | 12.58% | 1174 | 13.04% |
| 18 | 4919 | 12.15% | 1114 | 12.38% | 1072 | 11.91% |
| 19 | 4367 | 10.78% | 999 | 11.10% | 982 | 10.91% |
| 20 | 3983 | 9.83% | 891 | 9.90% | 875 | 9.72% |
| 21 | 3637 | 8.98% | 762 | 8.47% | 770 | 8.56% |
| 22 | 3187 | 7.87% | 714 | 7.93% | 715 | 7.94% |
| 23 | 2980 | 7.36% | 655 | 7.28% | 602 | 6.69% |
| 24 | 2627 | 6.49% | 576 | 6.40% | 590 | 6.56% |
| 25 | 2275 | 5.62% | 546 | 6.07% | 570 | 6.33% |
| 26 | 2162 | 5.34% | 467 | 5.19% | 504 | 5.60% |
| 27 | 1885 | 4.65% | 408 | 4.53% | 427 | 4.74% |
| 28 | 1776 | 4.39% | 385 | 4.28% | 396 | 4.40% |
| 29 | 1493 | 3.69% | 351 | 3.90% | 323 | 3.59% |
| Total | 40500 | 100.00% | 9000 | 100.00% | 9000 | 100.00% |

Based on the data in the above table, we have also given the illustration of answer length distribution ratio in different MRC datasets. For example, it can be seen from Figure 2(a). The blue squares represent the proportion of answers with length 4, the red squares represent the proportion of answers with length 5, and the green squares represent the proportion of answers with length 5.

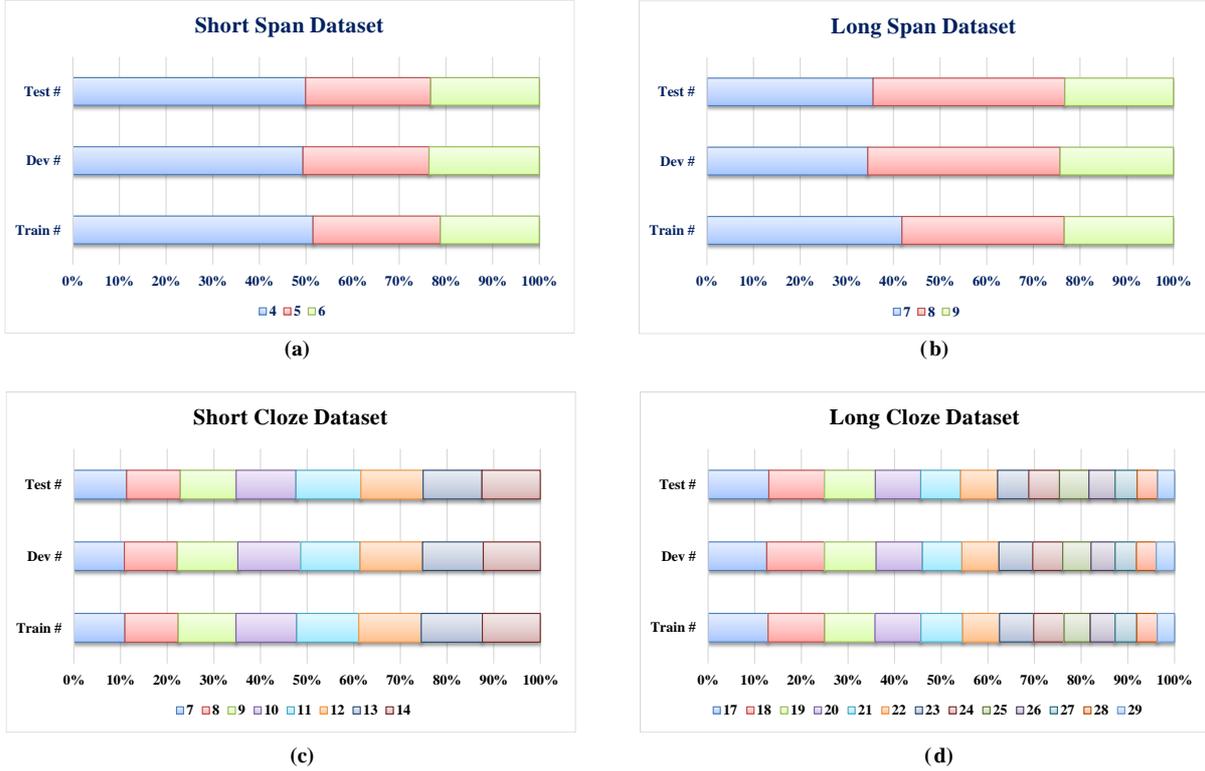

Figure 2: The illustration of answer length distribution ratio in different MRC datasets. (a) The short span dataset. (b) The long span dataset. (c) The short cloze dataset. (d) The long cloze dataset.

**Dataset Comparison**

The statistics of the proposed dataset have been given in the previous section. In this section, we compare the proposed dataset with the other MRC datasets. The comparison of the numbers of questions is shown in Table 7. In contrast to prior MRC datasets, the question size of proposed dataset is at a medium level.

Table 7: The number of questions of each MRC dataset

| Dataset | # Question | # Train Qu. | # Dev Qu. | # Test Qu. | Percentage of Train Set |
|---|---|---|---|---|---|
| SQuAD2.0 | 151,054 | 130,319 | 11,873 | 8,862 | 86.27% |
| SQuAD1.1 | 107,702 | 87,599 | 10,570 | 9,533 | 81.33% |
| TQA | 26,260 | 15,154 | 5,309 | 5,797 | 57.71% |
| MovieQA | 21,406 | 14,166 | 2,844 | 4,396 | 66.18% |
| MCScript | 13,939 | 9,731 | 1,411 | 2,797 | 69.81% |
| DREAM | 10,197 | 6,116 | 2,040 | 2,041 | 59.98% |
| OpenBookQA | 5,957 | 4,957 | 500 | 500 | 83.21% |
| ARC-E | 5,197 | 2,251 | 570 | 2,376 | 43.31% |
| WikiQA | 3,047 | 2,118 | 296 | 633 | 69.51% |
| ARC-C | 2,590 | 1,119 | 299 | 1,172 | 43.20% |
| ProPara | 488 | 391 | 54 | 43 | 80.12% |
| **Short Span** | **46,473** | **31,390** | **6,774** | **8,309** | **67.54%** |
| **Long Span** | **20,241** | **12,053** | **2,376** | **5,812** | **59.55%** |
| **Short Cloze** | **58,500** | **40,500** | **9,000** | **9,000** | **69.23%** |
| **Long Cloze** | **58,500** | **40,500** | **9,000** | **9,000** | **69.23%** |

Next, the statistics of the context size are given in Table 8. As we can see, in contrast to prior MRC datasets, the context size of proposed dataset is also at a medium level.

Table 8: The number of contexts of each MRC dataset

| Dataset | # Context | # Train Co. | # Dev Co. | # Test Co. | Unit of Context |
|---|---|---|---|---|---|
| Qangaroo-W | 51,318 | 43,738 | 5,129 | 2,451 | Passage |
| CoQA | 8,399 | 7,199 | 500 | 700 | Passage |
| CLOTH | 7,131 | 5,513 | 805 | 813 | Passage |
| DREAM | 6,444 | 3,869 | 1,288 | 1,287 | Dialogue |
| Qangaroo-M | 2,508 | 1,620 | 342 | 546 | Passage |
| TQA | 1,076 | 666 | 200 | 210 | Lesson |
| MovieQA | 548 | 362 | 77 | 109 | Movie |
| SQuAD1.1 | 536 | 442 | 48 | 46 | Article |
| SQuAD2.0 | 505 | 442 | 35 | 28 | Article |
| **Short Span** | **20,519** | **15,265** | **2,770** | **2,484** | **Passage** |
| **Long Span** | **10,484** | **7,756** | **1,163** | **1,565** | **Passage** |
| **Short Cloze** | **58,500** | **40,500** | **9,000** | **9,000** | **Passage** |
| **Long Cloze** | **58,500** | **40,500** | **9,000** | **9,000** | **Passage** |

We also compared the question style, answer style, source of corpora and generation method of each dataset. As shown in Table 9.

Table 9: Analysis of each MRC dataset

| Dataset | Question Style | Answer Style | Corpora Source | Generation Method |
|---|---|---|---|---|
| CliCR | Cloze | Free-Form | BMJ Case Reports | Automated |
| Facebook CBT | Cloze | Free-Form | Children's Books | Automated |
| DREAM | Natural | Free-Form | Exam | Crowdsourcing |
| MCScript | Natural | Free-Form | Narrative texts | Crowdsourcing |
| MovieQA | Natural | Free-Form | Movie Plot | Crowdsourcing |
| OpenBookQA | Natural | Free-Form | Elementary level science facts | Crowdsourcing |
| ProPara | Natural | Spans | Process paragraph | Crowdsourcing |
| Qangaroo-M | Synthesis | Spans | Wikipedia articles | Crowdsourcing |
| Qangaroo-W | Synthesis | Spans | Paper abstracts | Crowdsourcing |
| SQuAD1.1 | Natural | Spans | Wikipedia | Crowdsourcing |
| SQUAD2.0 | Natural | Spans | Wikipedia | Crowdsourcing |
| WikiQA | Natural | Free-Form | Relevant Wikipedia pages | Crowdsourcing |
| ARC-C | Natural | Free-Form | Grade-school science questions | Experts |
| ARC-E | Natural | Free-Form | Grade-school science questions | Experts |
| TQA | Natural | Free-Form | Middle school curricula | Experts |
| **Short Span** | **Natural** | **Spans** | **Wikipedia** | **Crowdsourcing** |
| **Long Span** | **Natural** | **Spans** | **Wikipedia** | **Crowdsourcing** |

| Short Cloze | Cloze | Multiple Choice | Wikipedia/News | Automated |
| Long Cloze | Cloze | Multiple Choice | Wikipedia/News | Automated |

## MLMs with Different Masking Lengths

In order to quantitatively verify whether masking schemes of different lengths will affect the performance of the MLM language model, in the previous section, we have proposed MRC tasks and constructed MRC datasets with different answer lengths. In this section, as shown in Figure 3, we use the above datasets and tasks to propose an evaluation framework for mask language models (MLMs) with different masking lengths. However, existing MLMs usually integrate various improvements. To eliminate the influence of the prior knowledge embedded in the existing MLMs, in this experiment, we do not directly use the existing MLMs, but conduct MLM training from scratch by ourselves. We trained four different MLMs, namely short span MLM, long span MLM, short cloze MLM and long cloze MLM. When training our MLMs, we used different masking lengths according to the average distribution of answer lengths in the proposed four MRC datasets.

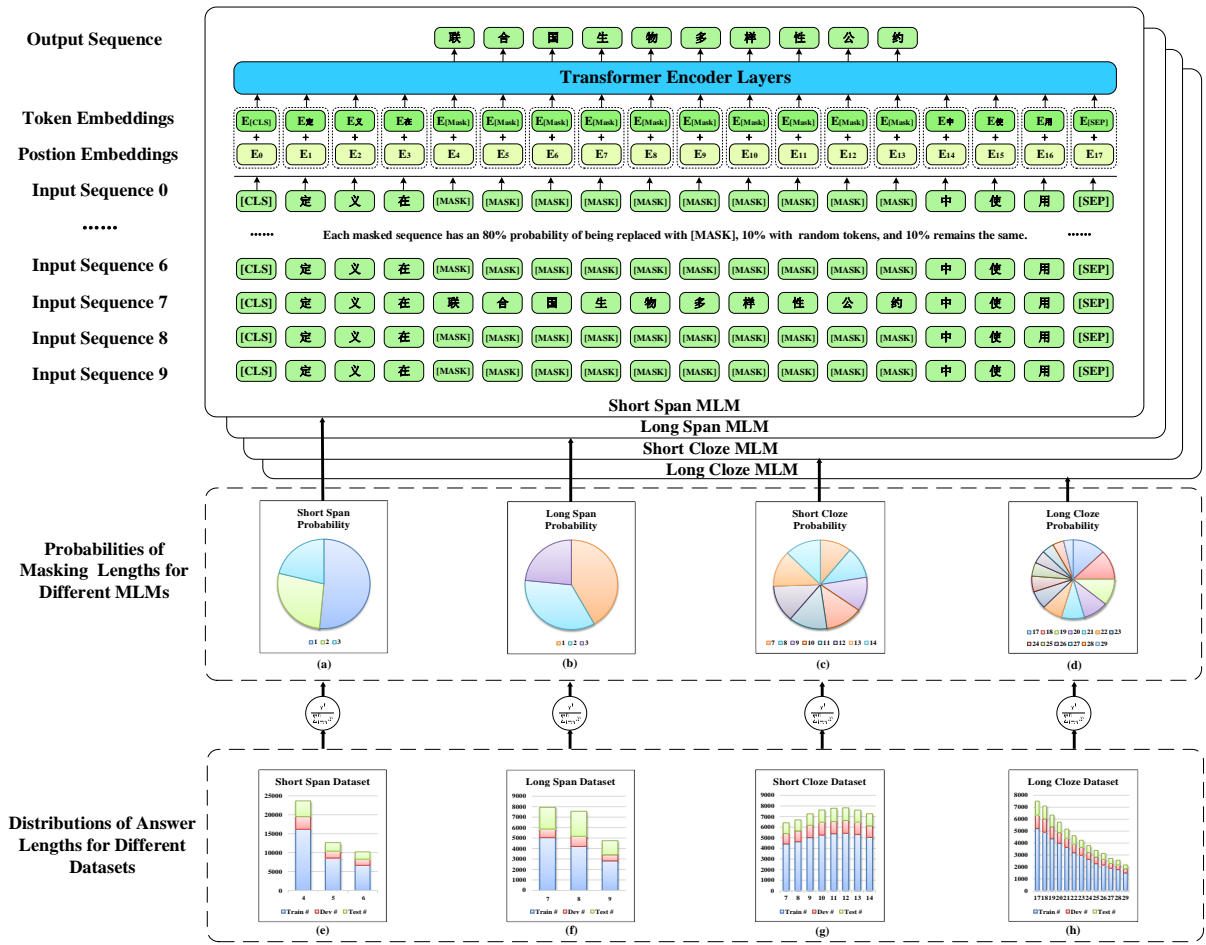

Figure 1: Illustration of the mask language models (MLMs) with different masking lengths according to the average distribution of answer lengths in the proposed four MRC datasets

**Masking Schemes**

The key point of our mask scheme used in our experiment is that the probability distribution of different masking lengths is equal to the proportional distribution of different answer lengths in the corresponding dataset. For example, as shown in Figure 3, in the short multiple-choice cloze dataset, the answer length distribution is shown in Figure 3 (e). Assuming that the total number of answers with length $l$ in the dataset is $X_l$, we can calculate the proportions of answers of different lengths in the dataset, which is $X_l / \sum_{l=min}^{max} X_l$. Then, we treat this proportional distribution as a probability distribution, and use it as the probability of different lengths being selected in the MLM. The pie chart of this probability distribution is shown in Figure 3(a).

In the MLM training process, first, we duplicate the input sequence 10 times, and then choose different ways to mask it. We use iterative sampling to mask the sequence. In each iteration, we will randomly select the current mask length according to the above probability distribution, such as $l$. Then, we randomly select a sequence with $l$ consecutive tokens from the paragraphs. This process will be cycled until the masking budget has been spent. Following BERT, masking budget is set to 15%, which means that 15% of the text in the paragraph will be selected.

Then, for each selected sequence, we also replace it with a proportion of 80% - 10% - 10%. As shown in Figure 3, the following span masking scheme in SpanBERT [9], we performs this replacement at the sequence level, rather than separately for each token, i.e., each selected sequence has an 80% probability of being replaced with "mask", 10% probability of being replaced with random tokens, and 10% remains the same.

We use the dynamic masking [14] to avoid masking the same sequences for each paragraph in every epoch. Following RoBERTa, we duplicate the input sequence 10 times, so that each sequence is masked in 10 different ways.

**Input Sequences**

Before feeding the training data into the model, we need to pre-process the data. A pre-processed input sample is a sequence composed of both the question and the reference context. A separation token (denoted as [SEP]) is used to separate question and context. It will be added between the question and the context, as well as the end of the context. In addition to [SEP], there are 4 special tokens in the input sequence:

**[CLS]:** Used to identify the beginning of the sequence. In tasks such as classification, it is usually necessary to use the output of the [CLS] position in the last layer.

**[UNK]:** The out-of-vocabulary (OOV) words will be replaced by this token.

**[PAD]:** Zero padding mask, for sentences shorter than the maximum length, we will have to fill [PAD] to make up for the length.

**[MASK]:** In some training objectives such as Masked Language Model (MLM), some input tokens are randomly replaced with [MASK] token (being masked) at and the model is required to predict the masked tokens.

After that, the question and context pair are tokenized. Commonly used tokenization methods are BERT tokenizer. In this baseline, we use "–vocab_path" to specify the Chinese vocabulary path. Then use this vocabulary to tokenize the question and context pair. Finally, each token is converted into a unique index according to the index of the corresponding Chinese character in the vocabulary.

**Tokenization**

We use the WordPiece tokenizer. The WordPiece tokenizer follows the subword tokenization scheme. The tokenizer first checks whether the word is in the vocabulary. If so, then it will be used as a token. If the word is not in the vocabulary, then the word will be split into subwords, and the tokenizer will constantly check that the split subword appears in the vocabulary after each split. Once a subword is found in the vocabulary, we use it as a token. The WordPiece tokenizer is very effective when dealing with out of vocabulary (OOV) words. Because there is no subword and no space between words in Chinese. We cannot apply WordPiece tokenizer to Chinese text directly. Thus, when tokenizing Chinese text with the WordPiece tokenization, following the Chinese BERT, we add spaces around all Chinese characters, and the input Chinese text will be split into Chinese characters, so all Chinese tokens (subwords) in the vocabulary are single Chinese characters.

**Embeddings**

In the embedding layer, the input indices are transformed into corresponding vector representation, which are usually obtained by adding three distinct representations, namely:

**Token Embeddings** (usually with shape (1, max length, hidden size)): Each input indices is transformed into a multi-dimensional word embeddings, which is randomly initialized from a standard Normal distribution with 0 mean and unit variance.

**Position Embeddings** (usually with shape (1, max length, hidden size)): It is used to indicate the position of the token, which is a learned embedding vector. This is different from normal transformer in BERT, which has a pre-set value.

Finally, these embeddings are summed element-wise to produce a single vector representation and fed into the transformer encoders.

## Experiments

**Pre-training Setup**

Using the open source framework of UER-py [UER-py], we pre-train MLM models with different lengths on Chinese corpus. Compared with the original BERT implementation, the main points in our implementation include:

(a) The probability distribution of different masking lengths is equal to the proportional distribution of different answer lengths in the corresponding dataset;

(b) We did not use the next sentence prediction (NSP) training objective, but only masked language model (MLM) ;

(c) We perform the mask replacement at the sequence level instead of performing this replacement separately for each token;

(d) We use the dynamic masking instead of static masking;

Our implementation of MLM is trained on the cloud with Nvidia V100 GPU. We use a sequence of up to 512 tokens for pre-training. The learning rate is set to 5e-5 and batch size is 64. The parameters of Adam optimizer are fixed at $β1=0.8$, $β2=0.9$. We use the mixed precision training to reduce the memory usage and accelerate the pre-training.

**Training Corpus**

The effective pre-training of the MLM crucially relies on large-scale training data from various domains. Improving the diversity of data domains and increasing the amount of data can result in the improved performance in downstream tasks. [9,32]

In this work, we collect the Chinese training data from the Internet, and then cleaned the data. For example: clean the html mark, remove extra empty characters, remove the picture mark, etc. Finally, we collect a Chinese corpus for our MLM pre-training. The corpus contains Chinese texts in the following domains:

(1) Wikipedia, contains various Chinese Wikipedia documents.

(2) Chinese academic papers on the Wan Fang database.

(3) Chinese social text messages on the Weibo.

(4) Chinese articles on the WeChat official accounts.

(5) Chinese news articles, including titles, keywords, descriptions, and text. Categories include: finance, real estate, stocks, home furnishing, education, technology, society, fashion, current affairs, sports, constellations, games, entertainment.

**Fine-tune and Evaluation**

We fine-tune and evaluate our pre-trained masked language models on four machine reading comprehension datasets, namely short span extraction dataset, long span extraction dataset, short multi-choices cloze dataset, and long multi-choices cloze dataset, and the details are given below.

**Span Extraction**

Span extraction reading comprehension task is composed of passages, questions and answers. This task requires the computers to answer relevant questions according to the passages. The answer to the question can be found in the passage, that is, the answer is a span (fragment) in the passage.

This task can be simplified to predict a starting position and an ending position in a passage, and the answer is a text span between the start and the end.

The process of using MLM to deal with span extraction reading comprehension tasks can be divided into three layers: input layer, transformer based encoder layer and output layer.

(1) Input layer

In the input layer, we preprocess the input passage and questions, first we perform word piece tokenization, then splice the questions and passages, we insert [CLS] at the beginning of the input sequence, and [SEP] at the end and the dividing point between the question and the passage, so as to finally get the input sequence.

$$S = [CLS], q_1, \ldots, q_m [SEP], p_1, \ldots p_n, [SEP] \qquad (9)$$

It should be noted that if the length n of the input text is less than the maximum sequence length N, the padding token [PAD] needs to be spliced after the input sequence until it reaches the maximum sequence length n. For example, in the following example, assume that the maximum sequence length of our model is n = 10 and the current input sequence length is 7. Then three padding tokens [PAD] are required after the input sequence.

$$S = [CLS], q_1, \ldots, q_m [SEP], p_1, \ldots p_n, [SEP] \ [PAD][PAD][PAD] \qquad (10)$$

Conversely, if the length of current input sequence is longer than the maximum sequence length N, the sequence needs to be sliced and divided into multiple sub-sequences. For example, assume that the maximum sequence length of our model is n = 10 and the current input sequence length is 40. Then the model can only process input sequences with a length of 10 tokens at one time and the sequence needs to be divided into 4 sub-sequences.

In addition, it should be noted that we have to put the question at the beginning of the input sequence. Because if the question is divided into multiple sub-sequences, the question cannot be answered. If the passage is divided into multiple sub-sequences, the answers in the passage can be obtained through other sequences.

(2) Encoder layer based on the transformers

The input sequence will be converted into the token embeddings, position embeddings, and segment embeddings. These three embeddings will be added to obtain the input vector. The input vector will pass through 12 encoding layers. In these encoding layers, with the help of multi-head self-attention mechanism, the model will fully learn the semantic association between passages and questions.

(3) Output layer

The output of the last layer of the transformer encoder passes through a full connection layer, and predicts the probability PS of each position as the answer and the probability PE of the end position through Softmax.

Then, we input the prediction probabilities and the ground truth positions into the cross entropy loss function at the same time to obtain the loss of the model. Finally, the cross entropy loss at the starting position and the loss at the ending position are averaged to obtain the final total cross entropy loss of the model. The training objective is to minimize the total cross entropy loss between the prediction probability and the ground truth position.

(4) Answer prediction and evaluation

In the output layer, we select the starting position and ending position with the highest probability as the prediction answer. Finally, F1 and EM of the predicted answer are calculated according to the standard answer.

**Multi-choices Cloze**

In sentence cloze-style reading comprehension task, we select several sentences in the passages and replace with special marks (for example, [BLANK]) to form an incomplete passage. The selected sentences will form the candidate list, and the computer is required to fill in the blanks with the right candidate sentences.

(1) Input sequence

The input sequence is composed of an answer option and the passage (with blanks), and then the semantic representation of the context is obtained through transformer encoder layers. Finally, the probability of each blank corresponding to an option is output. It should be noted that the two components of the input sequence are an answer option and an incomplete passage with multiple blanks. Because there are 9 blanks (corresponding to 9 different options) in each passage in our dataset. Therefore, we need to enter 9 different sequences, and each sequence contains an option.

For example, assume that the current answer options are: $a_1, \ldots a_i, \ldots, a_m$, where $a_i$ represents the *i*-th word in the answer option text. Let the input paragraph with blanks be $p_1, \ldots blank_1, \ldots blank_i \ldots, p_n$ where $p_n$ represents the *n*-th word of the input passage and $blank_i$ represents the *i*-th masked answer. The input sequence $S$ can be expressed as:

$$S = [CLS], a_1, \ldots, a_m [SEP], p_1, \ldots blank_1, \ldots blank_i \ldots, p_n, [SEP] \tag{11}$$

Where [CLS] represents the special token at the beginning of the input sequence, [SEP] represents the segmentation token and the end token of the input sequence (following BERT).

It should be noted that if the length n of the input sequence is less than the maximum sequence length *N*, the padding token [PAD] needs to be spliced after the input sequence until it reaches the maximum sequence length *N*. For example, in the following example, assume that the maximum sequence length is n = 10 and the input sequence length is 9. Then one padding token [PAD] are required after the input sequence.

$$S = [CLS], a_1, \ldots, a_m [SEP], p_1, \ldots blank_1, \ldots blank_i \ldots, p_n, [SEP][PAD] \tag{12}$$

Conversely, if the input sequence length n is longer than the maximum length *N*, it needs to be truncated into multiple input sequences. Here, we usually put the answer option in the front, so that the answer options will not be truncated.

(3) Embeddings

This section describes how to preprocess the input sequence to get the corresponding input representation. The input representation is composed of the sum of a token embedding, segment embedding and position embedding. For example, assume these three embeddings are $E_{token}$, $E_{segment}$ and $E_{pposition}$ respectively, the input representation $E$ corresponding to the input sequence can be calculated by the following formula:

$$E = E_{token} + E_{segment} + E_{position} \tag{12}$$

In the formula, $E_{token}$ represents the token embedding, $E_{segment}$ represents the segment embedding, and $E_{position}$ represents the position embedding, the size of the three embeddings are all *M*d*, and *M* represents the maximum length of the sequence, which is 512 in this paper, and d represents the dimension of the word vector, which is 768 in this article.

(4) Transformer Encoders

In transformer encoders, the input embeddings pass through 12 encoder layers, and uses the self-attention mechanism to fully learn the semantic representation between each word in the input sequence.

$$E_i = Encoder\_layer(E_{i-1}) \tag{13}$$

Where $E_{i-1}$ indicates the output vector of the *i*-th encoder layer, and $E_0$ is specified to be equal to the input embedding. Finally, after 12 encoder layers, the output vector of the encoder is:

$$E_o = Encoder\_layer(E_{11}) \tag{14}$$

Where $E_o$ indicates the output vector of the last encoder layer.

(5) The pooler layer

In this layer, the output of the encoder layer is fed into a pooler layer to get pooled output.

$$Pooled\_output = E_o * W + b \tag{15}$$

Where $W$ denotes the weight matrix of the pooling layer and $b$ represents the bias vector.

(6) The output layer

In the final output result, we don't need the output of each token in the input sequence, but only need the output sequence where the current blank is located. Therefore, for the pooled outputs of the other positions except blanks, we will remove them from the total output, and then splice the remaining pooled outputs of these blanks positions to obtain the output $O$. Among them, for the output $O_i$ denoting the i-th blank, we use the softmax function to calculate the confidence probability that the current blank position matches the current option.

$$P_i = \text{Softmax}(O_i) = \frac{e^{O_i}}{\sum_j e^{O_j}} \tag{16}$$

Finally, after obtaining the prediction probability $Pi$ corresponding to the class label of the current sequence, the cross entropy loss between the correct answer $T_i$ and the prediction probability $P_i$ is calculated.

$$Loss = \text{CrossEntropyLoss}(P_i, T_i) \tag{17}$$

The training objective is to minimize the total cross entropy loss between the prediction probability and the standard answer sequence.

(7) Answer prediction and evaluation

When predicting the answer, we choose the blank with the highest probability as the position where the answer option I should be filled. Finally, according to the standard answer, the PAC and QAC of the predicted answer are calculated.

## Result Analysis

### Human Performance

In order to evaluate the human performance on our datasets, we invited 10 college students to answer questions in the datasets manually. Finally, we got the answers in the test sets of the four datasets respectively. Then we calculated F1 and EM to roughly evaluate the human performance on the proposed long span dataset and short span dataset, and we also calculated PAC and QAC on the proposed cloze datasets.

### Model Performance

The evaluation results on the pre-trained models on different MRC datasets are presented in Table 10. For fair comparison, these models are all fine-tuned with the same hyper-parameters and without any data augmentation. We fine-tuned three different runs and report the mean results. The pre-trained language models based on the corresponding MLMs constantly outperform other pre-trained language models on the corresponding datasets by an obvious margin.

As shown in Table 10, the pre-trained language model based on the long span MLMs performs better on the long span dataset compared to other three pre-trained language models, though there still exists a large gap between this model and human performance. At the same time, the pre-trained language model based on the short span MLMs performs better on the short span dataset compared to other pre-trained language models.

Table 10: The evaluation results of pre-trained models and human performance on span datasets.

| Models | Short Span Dataset | | Long Span Dataset | |
|---|---|---|---|---|
| | F1 | EM | F1 | EM |
| **Human Performance** | 93.56 | 85.34 | 90.47 | 82.56 |
| PLM with Short Span MLM | **25.03** | **23.93** | 13.09 | 10.79 |
| PLM with Long Span MLM | 24.74 | 23.31 | **13.28** | **11.06** |

As shown in Table 11, pre-trained with long cloze MLMs, the long pre-trained model outperforms other models on the long cloze dataset. As for short cloze dataset, the pre-trained language model based on the short cloze MLMs achieves score increase over other models, demonstrating the effectiveness of the proposed MLMs.

Table 11: The evaluation results of pre-trained models and human performance on cloze datasets.

| Models | Short Cloze Dataset | | Long Cloze Dataset | |
|---|---|---|---|---|
| | QAC | PAC | QAC | PAC |
| **Human Performance** | **97.45** | **92.87** | **95.24** | **90.71** |
| PLM with Short Cloze MLM | **64.63** | **17.10** | 65.82 | 17.60 |
| PLM with Long Cloze MLM | 63.28 | 15.20 | **67.02** | **18.60** |

The experimental results demonstrate that our hypothesis is true. The length of MLM is indeed related to their performance in MRC tasks with different answer lengths. It can guide us how to pre-train an MLM model with a relatively suitable mask length distribution for various MRC tasks.

## Conclusions

In this paper, we propose an evaluation framework to quantitatively verify whether masking schemes of different lengths will affect the results of the MLM language model in MRC tasks with different answer length. In order to address this issue, herein, (1) we propose four MRC tasks with different answer length distributions, namely short span extraction task, long span extraction task, short multiple-choice cloze task, long multiple-choice cloze task; (2) four Chinese MRC datasets are created for these tasks; (3) we also have pre-trained four masked language models according to the answer length distributions of these datasets; (4) ablation experiments are conducted on the datasets to verify our hypothesis. The experimental results demonstrate that our hypothesis is true. It can guide us how to pre-train an MLM model with a relatively suitable mask length distribution for various MRC tasks. However, as a case study, we must also be conservative in the strength of our conclusions since more comprehensive future research and experiments are needed.

## Conflicts of Interest

The authors declare no conflict of interest.

## Funding Statement

Research reported in this publication was partially supported by the Ministry of science and technology of China under the project of "New Generation Artificial Intelligence" with No.2018AAA0101803, and the Science and Technology Project of Guizhou Province with No.[2015] 4011. It was also partially funded by Science and Technology Project of Guizhou Province with No.[2017] 5788.